\documentclass{article}
\usepackage{iclr2027_conference,times}

\usepackage{amsmath,amsfonts,bm}

\def\eqref#1{equation~\ref{#1}}

\def\1{\bm{1}}

\DeclareMathAlphabet{\mathsfit}{\encodingdefault}{\sfdefault}{m}{sl}
\SetMathAlphabet{\mathsfit}{bold}{\encodingdefault}{\sfdefault}{bx}{n}

\usepackage{fontawesome5}
\usepackage{fancyhdr}
\usepackage{hyperref}
\usepackage{url}
\usepackage{amsmath}
\usepackage{amssymb}
\usepackage{booktabs}
\usepackage{array}
\usepackage{xcolor}
\usepackage[most]{tcolorbox}
\usepackage{enumitem}
\usepackage{tikz}
\usepackage{graphicx}
\usetikzlibrary{arrows.meta,positioning}
\newcolumntype{P}[1]{>{\raggedright\arraybackslash}p{#1}}
\newcommand{\casecorrect}{\colorbox{green!15}{\textcolor{green!45!black}{\textbf{Correct}}}}
\newcommand{\caseincorrect}{\colorbox{red!12}{\textcolor{red!65!black}{\textbf{Incorrect}}}}
\newtcblisting{promptbox}[2][blue]{%
    enhanced,
    breakable,
    listing only,
    colback=#1!3!white,
    colframe=#1!70!black,
    colbacktitle=#1!70!black,
    coltitle=white,
    fonttitle=\bfseries,
    title={#2},
    left=5pt,
    right=5pt,
    top=4pt,
    bottom=4pt,
    boxsep=0pt,
    arc=2pt,
    listing options={basicstyle=\ttfamily\scriptsize,breaklines=true,columns=fullflexible,keepspaces=true}
}

\title{EnSIMem: Entity-Structured Indexing for Long-Term Agent Memory}

\author{
\textbf{Xuanyu Meng}$^{1}$, \textbf{Xing Fan}$^{2}$, \textbf{Xinyi Fan}$^{1}$,
\textbf{Chenlei Guo}$^{2}$, \textbf{Yixuan Xie}$^{1}$, \textbf{Jiawei Han}$^{1}$\\[2pt]
\normalfont
$^{1}$University of Illinois Urbana-Champaign\\
$^{2}$Amazon\\[2pt]
\texttt{\{xuanyum2, xfan31, xie39, hanj\}@illinois.edu}\\
\texttt{\{fanxing, guochenl\}@amazon.com}}

\iclrfinalcopy
\begin{document}

\maketitle
\fancyhead{}

\begin{abstract}
An agent that interacts with users over long periods must recall facts, preferences, events, and changes from a continuously growing interaction history. Existing memory systems often compress interactions into generic summaries or retrieve anonymous text chunks, making it difficult for an agent to identify the correct entity, property, and supporting evidence. We present EnSIMem, an entity-structured long-term memory architecture for an agent. During offline construction, the system organizes interactions into theme-coherent episodes and builds dialogue-grounded index entries of the form $[\textit{entity}][\textit{entity\_type}][\textit{property}:\textit{value}]$. Each entry preserves its source turns, temporal information, and available multimodal fields. During online interaction, the agent's request is decomposed into evidence requirements whose properties are aligned with the memory index. Entity-property lookup and adaptive retrieval then collect the evidence needed for point, temporal, compositional, and aggregation reasoning. The agent generates its response from the preserved source evidence rather than from lossy memory summaries. On long-term agent-memory benchmarks, EnSIMem achieves high answer accuracy while maintaining compact, evidence-focused contexts. These results show that entity-structured indexing and episode-level provenance provide a reliable foundation for long-term memory in agents. The code of our model is available at
\faGithub\ 
\href{https://github.com/RamonMeng/EnSIMem}
{\nolinkurl{https://github.com/RamonMeng/EnSIMem}}.
\end{abstract}

\section{Introduction}

An agent deployed over a \textit{long term} must accumulate and reuse information from its past interactions. This information may include user preferences, personal facts, plans, commitments, events, feedback, and decisions that influence future actions. For an agent, memory is therefore not simply a larger context window: it is a persistent interface between past experience and current reasoning, as reflected by memory-stream, tiered-memory, and evolving-memory architectures \citep{park2023generative,packer2023memgpt,xu2025amem,kang2025memoryos}. The memory system must help the agent recover the right evidence at the right time while preserving enough context for the agent to interpret that evidence correctly.

Long-term agent memory is particularly challenging because relevant information is distributed across many interactions and may be expressed in different ways. This challenge is central to long-term conversational-memory benchmarks and interactive memory evaluations \citep{maharana2024locomo,wu2024longmemeval}. A user may describe an event with a concrete action while a later request refers to the same event using a more general expression. Important information may also be introduced through pronouns, temporal references, correlations, or multimodal observations. For example, an interaction may state that a person traveled to Chicago by helicopter, while a later request asks how that person traveled to Chicago. A useful memory system must connect these expressions without collapsing the underlying event into an uninformative category such as ``activity'' or ``travel.'' It must also retain the original evidence so that the agent can distinguish an explicit fact from an inference.

Existing approaches expose a tension between scalability and specificity. Providing the complete interaction history gives the agent access to all evidence, but causes context growth, higher latency, and distraction effects \citep{liu2024lost,du2025context,wang2024loong,bai2025longbenchv2}. Summarization reduces the context size but can remove details, temporal qualifiers, and provenance. Conventional retrieval-augmented generation systems retrieve anonymous text chunks or vector-nearest memories \citep{lewis2020rag,karpukhin2020dpr,izacard2022contriever,khattab2020colbert}; more recent systems use iterative or structured retrieval to improve multi-step reasoning \citep{asai2024selfrag,trivedi2023ircot,edge2024graphrag}. These representations are compact, but they do not explicitly indicate which entity, property, or event makes a memory relevant. EnSIMem addresses this tension by using structure to identify the right evidence and the original dialogue to interpret it: a large interaction history is converted into a small, query-specific, evidence-rich buffer instead of being passed wholesale to the agent.

\begin{figure*}[t]
    \centering
    \makebox[\textwidth][c]{%
        \includegraphics[
            width=1.05\textwidth,
            trim=10 6 10 6,
            clip
        ]{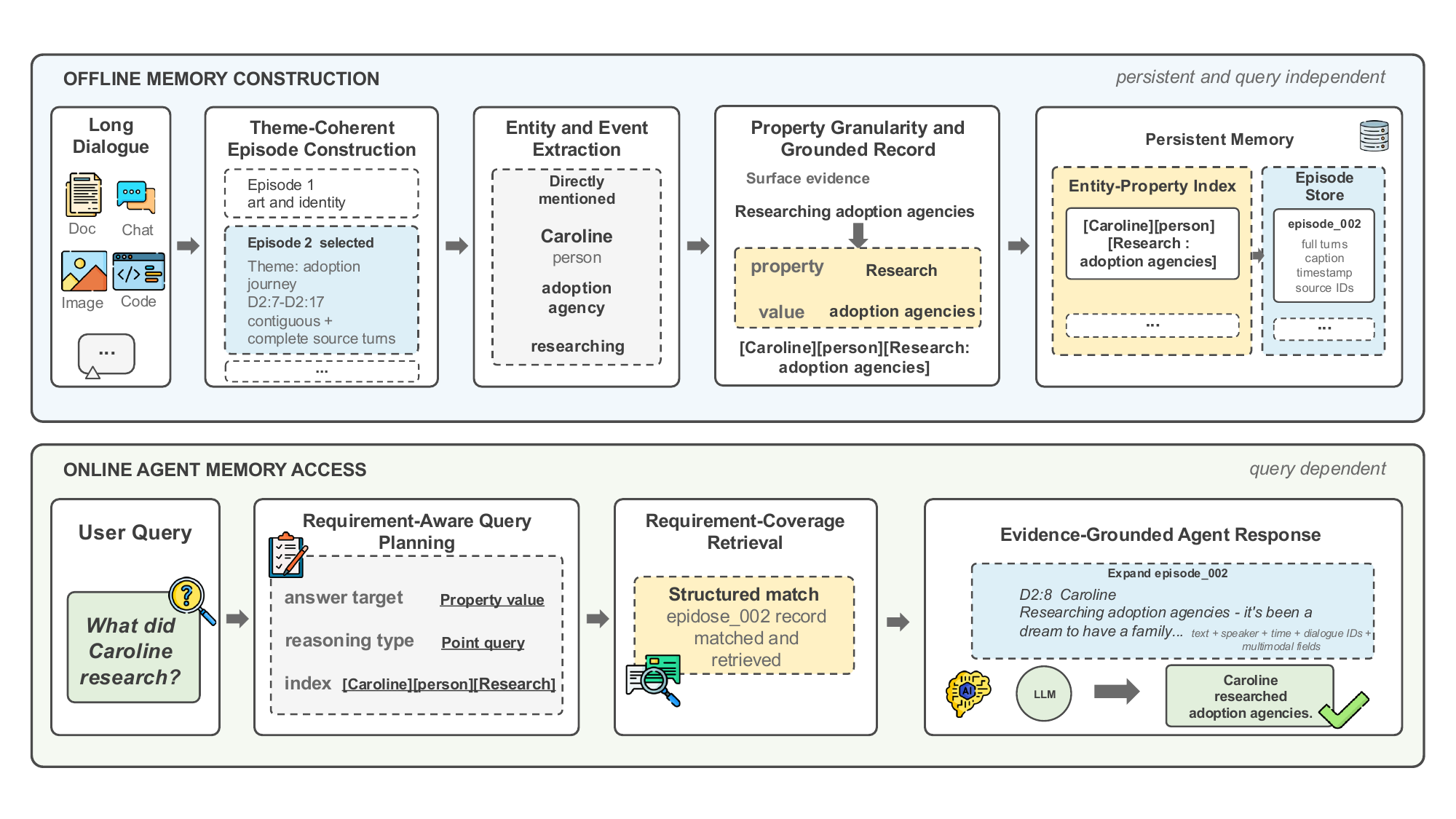}%
    }
    \caption{
    Overview of EnSIMem. Offline memory construction preserves multimodal dialogue, partitions it into theme-coherent episodes, extracts dialogue-grounded entity-property records, and stores an entity-property index linked to the original episodes. Online, the agent decomposes the request, performs requirement-aware retrieval with structured matching, expands the selected episode into source evidence, and generates a grounded response.
    }
    \label{fig:ensi-memory-overview}
\end{figure*}

We present EnSIMem, an entity-structure indexing-based long-term memory architecture for agentic systems, building on entity-structured retrieval and structure-augmented reasoning ideas \citep{meng2026ensirag,parekh2025structure}.
Figure~\ref{fig:ensi-memory-overview} summarizes the overall EnSIMem architecture. The central design principle is to use structure as an address system for memory, rather than as a lossy replacement for the original interaction. During offline construction, EnSIMem organizes conversations into theme-coherent episodes and builds dialogue-grounded index records of the form
$[\textit{entity}][\textit{entity\_type}][\textit{property}:\textit{value}]$.
The entity may be a person, object, or salient event. The property is selected at an intermediate level of granularity: broad enough to connect paraphrases, but specific enough to preserve the identity of the underlying relation or event. When available, a finer-grained value records the concrete realization of that property. Every record remains linked to its source turns, timestamps, and multimodal fields.

Upon interaction, the query is decomposed into explicit evidence requirements. This follows the broader use of iterative query decomposition and agentic query rewriting for complex retrieval tasks \citep{trivedi2023ircot,shankar2024docetl}. The query planner extracts entities and properties at the same granularity as the memory index and identifies whether the query requires point evidence, temporal comparison, compositional reasoning, or aggregation. Structured entity-property lookup is combined with dense fallback retrieval when the wording differs from the indexed property. Retrieval then proceeds adaptively until the requirements are sufficiently covered. This allows the system to retrieve a small candidate set for a point query while preserving exhaustive coverage for a query that requires counting or comparing multiple events.

The final response is generated from the retrieved source evidence rather than from index entries alone, consistent with evidence-grounded graph and structure-aware generation approaches \citep{he2024gretriever,hu2024grag}. This separation gives the agent both efficient access and evidentiary grounding: the index identifies where to look, while the original episode provides the context needed for interpretation and answer synthesis. It also makes the memory pipeline inspectable. Each answer can be traced from a query requirement, through an entity-property access path, to the dialogue turns and multimodal observations that support it.

EnSIMem separates query-independent memory construction from query-dependent recall. The offline representation can therefore be reused across many future queries, while the online stage spends computation only on the evidence required by the current query. On long-term conversational memory benchmarks, EnSIMem achieves high answer accuracy with compact, evidence-focused retrieval contexts. The results suggest that theme-coherent episodes, dialogue-grounded surface-property indexing, and requirement-aware retrieval provide a practical foundation for reliable long-term memory in agents.

Our contributions are:
\begin{enumerate}[leftmargin=*,itemsep=2pt,topsep=2pt]
    \item \textbf{Theme-coherent episodic memory construction.}
    We organize long interactions into theme-coherent, contiguous episodes that preserve local context, temporal order, and provenance.

    \item \textbf{Dialogue-grounded entity-property indexing.}
    We index originally mentioned entities and salient events with structured records linked to their original dialogue turns and multimodal observations.

    \item \textbf{Granularity-controlled property alignment.}
    We align corpus and query properties at an intermediate level that supports paraphrase matching without erasing event-specific distinctions.

    \item \textbf{Requirement-aware evidence localization.}
    We decompose agent requests into explicit evidence requirements and retrieve memories through structured entity-property matching.

    \item \textbf{Query-type-aware adaptive retrieval.}
    We distinguish point queries from temporal, compositional, and aggregation queries and allocate evidence budgets appropriate to their reasoning requirements.

    \item \textbf{An evidence-rich short-context buffer for agent response generation.}
    We convert long interaction histories into compact, query-specific reasoning buffers that retain relevant source evidence while filtering unrelated noise, without additional training or supervision.  This methodology leads to state-of-the-art results, as shown in our experiments.
      
\end{enumerate}

\section{Related Work}

\paragraph{Memory in long-running agents.}
Agent memory can be viewed through three complementary functions. Episodic memory preserves situated experiences and their temporal context; memory streams, reflection, tiered context management, and continual personalization provide representative designs \citep{park2023generative,packer2023memgpt,zhong2023memorybank}. Semantic memory consolidates relatively stable facts and relations, as in temporal graphs, persistent fact stores, evolving notes, and hierarchical memory systems \citep{rasmussen2025zep,chhikara2025mem0,xu2025amem,kang2025memoryos}. Procedural memory supports reusable behavior through feedback, action traces, or acquired skills \citep{shinn2023reflexion,wang2023voyager}. Recent surveys organize these systems by memory substrate, cognitive function, and memory subject, and emphasize that storage, retrieval, updating, and forgetting are increasingly treated as learnable agent operations \citep{huang2026agentmemorysurvey}. These functions may coexist, but they impose different update and evaluation requirements. This work is deliberately confined to episodic conversational memory: EnSIMem organizes past interactions for evidence-grounded recall, while extending the same address-and-provenance principle to semantic and procedural memory remains a promising direction for future work.

\paragraph{Recent agent-memory systems and evaluations.}
Recent work increasingly treats memory as an agent capability rather than a fixed summary. PlugMem builds a task-agnostic memory graph, AgeMem learns long- and short-term memory operations as policy actions, and Nemori adaptively distills experiences \citep{yang2026plugmem,yu2026agemem,ma2026nemori}. New evaluations include multimodal conversations, while MemAudit introduces MEMPROBE for recovering hidden, evolving user state \citep{bei2026memgallery,ma2026memprobe}; procedural-memory work studies reusable skills and their maintenance \citep{belikova2026procedural,song2026skillops,xu2026agentskills}. ReasoningBank and SkillOS further study reusable reasoning traces and skill curation for self-evolving agents \citep{ouyang2026reasoningbank,ouyang2026skillos}. A recent survey organizes this frontier by memory substrate, function, and subject \citep{huang2026agentmemorysurvey}. EnSIMem complements these efforts by preserving original episodic evidence and using structure only to address it, rather than replacing it with a generalized artifact.

\paragraph{Entity-structured retrieval for long documents.}
Dense and graph-aware retrieval support general-purpose and multi-hop access \citep{lewis2020rag,karpukhin2020dpr,khattab2020colbert,edge2024graphrag,gutierrez2024hipporag}, while Loong, Oolong, and SLIDERS study context selection and structured reasoning over long documents \citep{wang2024loong,bertsch2025oolong,joshi2026sliders}. Their induced structures do not target a persistent, dialogue-grounded memory with temporal and multimodal provenance. EnSI-RAG introduces entity-structure indexing for long-document QA \citep{meng2026ensirag}; EnSIMem extends this idea to agent memory by linking records to episodes and query-aware evidence budgets.

\paragraph{Structured agent memory.}
The baselines in Tables 1 and 2 cover the main choices: Full Context, RAG, HippoRAG, Zep, Mem0, LangMem, Nemori, and MEMORA \citep{lewis2020rag,gutierrez2024hipporag,rasmussen2025zep,chhikara2025mem0,langchain2025langmem,ma2026nemori,xia2026memora}. They improve compactness or navigation, but may lose local wording, provenance, or a direct address to the requested entity-property pair. EnSIMem instead uses structure as an address into preserved episodes, so the answer model reasons over original turns. Detailed benchmark definitions and baseline specification can be found in Appendix~\ref{app:benchmarks}.

\section{EnSIMem}

\subsection{System Objective and Design Principle}

Given an interaction history $\mathcal{D}$ and a new request $q$, EnSIMem must locate the relevant past evidence without exposing the agent to the full growing transcript. We separate reusable, query-independent memory construction from query-dependent access:
\begin{equation}
\mathcal{M}=F_{\mathrm{memory}}(\mathcal{D}),
\qquad
\hat{y}=G\!\left(q,R\!\left(q,\mathcal{M}\right)\right),
\end{equation}
where $\mathcal{M}$ is persistent memory, $R$ retrieves evidence for $q$, and $G$ generates the answer $\hat{y}$. The central design choice is that the index is an \emph{address system}, not a compressed substitute for the interaction: structure determines where to look, while original evidence determines what to answer.

This distinction addresses two different failure modes. Searching the complete transcript preserves information but leaves the answer model to separate a small amount of useful evidence from extensive distractors. Reasoning directly from compressed memories is cheaper, but a summary may omit the qualifier, timestamp, or wording that makes an answer correct. EnSIMem instead compresses the \emph{access path}: the structured record is deliberately small, but following it restores the original evidence before reasoning. The long-context problem is thereby converted into a short-context problem without treating the short, structured representation as the evidence itself.

\subsection{Offline Episodic Memory Construction}

EnSIMem partitions each session into contiguous, theme-coherent episodes. Neighboring turns remain together while they concern the same entity, event, goal, or unresolved conversational reference; a boundary is introduced only after a sustained topic or task shift. This middle-level granularity retains pronouns and local context that per-turn units can lose, while avoiding the unrelated material admitted by whole-session units. The partition is unsupervised and need not recover one uniquely correct boundary: what matters is preserving the same evidence at a comparable semantic scale, a claim tested by the episode-granularity ablation.

For every episode $e$, EnSIMem extracts explicitly mentioned entities and salient events and represents their supported properties as
\begin{equation}
r=\langle x,t,\mathcal{P},e,\pi\rangle,
\qquad
\mathcal{P}=\{(p_i,v_i)\}_{i=1}^{m},
\end{equation}
where $x$ is an entity, $t$ its type, $m$ is the number of properties extracted from episode $e$, and each $(p_i,v_i)$ is a supported property-value pair. The link $\pi$ points to the turns that support the record in $e$. The retrieval-facing view is therefore
\begin{equation}
[x][t][\{p_i:v_i\}_{i=1}^{m}]
\longrightarrow
\mathit{episode\_id}.
\end{equation}
Properties are normalized only to an intermediate semantic level: broad enough to align paraphrases, but specific enough to distinguish relations and events. For example, ``took a helicopter to Chicago'' supports two pairs, $\{\mathit{destination}:\mathit{Chicago},\mathit{means\_of\_transport}:\mathit{helicopter}\}$, whereas ``traveled to Chicago'' supports only $\{\mathit{destination}:\mathit{Chicago}\}$. The set-valued representation aligns both phrases on their shared destination while preserving the additional transport information only when it is explicitly stated. Unrelated actions are not collapsed into a universal \emph{activity} label. This compatible granularity lets corpus and query properties align without a fixed dataset ontology, while the episode link preserves temporal, conversational, and multimodal context.

Episodes and index records play complementary roles. An episode supplies a coherent evidence unit: it keeps together nearby turns needed to resolve pronouns, ellipsis, and temporal references. The index exposes only the entities and relations needed to find that unit. Neither representation is sufficient alone---episodes without addresses remain difficult to locate, while records without episodes lose the context required to interpret them. Their combination permits aggressive filtering at retrieval time without discarding the information needed at answer time.

\subsection{Online Requirement-Aware Retrieval}

For a request $q$, the planner produces atomic evidence requirements
\begin{equation}
H(q)=\{h_1,\ldots,h_K\},
\qquad
h_i=(x_i,t_i,p_i,v_i,c_i),
\end{equation}
where each requirement specifies a target entity $x_i$, type $t_i$, compatible property $p_i$, and optional value or condition. The planner also identifies whether the request is point, temporal, compositional, or aggregative. This classification sets the evidence budget: a point question may stop after one supported requirement, whereas counting or comparison continues until all necessary episodes are covered.

Structured matching first finds index records compatible with $H(q)$ and follows them to source episodes. Dense retrieval over the textual index representation protects recall when wording differs or a structured field is incomplete. If $I$ is the structured index and $T$ its textual form, the candidate set is
\begin{equation}
\mathcal{E}(q)=\operatorname{Dedup}\!\left(
\operatorname{Structured}(I,H(q))
\;\cup\;
\operatorname{Dense}(T,H(q))
\right).
\end{equation}
Candidates are merged by episode identity, and retrieval stops according to requirement coverage rather than a fixed similarity threshold or universal top-$k$. Structured access provides precise addresses; dense fallback and adaptive continuation protect recall.

Requirement coverage is important because relevance is not the same as sufficiency. A highly similar episode may answer a point question, yet the same stopping rule can fail for a request that asks for all preferences, compares two dates, or composes facts about several entities. The planner therefore determines what evidence must be present before retrieval begins, and the retriever tests coverage against that plan. This changes retrieval from selecting the most similar memories to assembling the smallest evidence set that supports the requested reasoning operation.

The resulting access path is reusable because the expensive organization is independent of any particular future question. Once an episode has been assigned a theme and its records have been linked to source turns, many requests can use the same address without rebuilding the memory. The online planner only changes the evidence requirements and stopping condition. This separation also makes the behavior inspectable: a reviewer can identify which requirement selected an episode, which record supplied the address, and which original turns supported the answer. In this sense, EnSIMem does not assume that a single summary is sufficient for every downstream task; it stores a compact index for navigation and defers interpretation until the relevant evidence has been restored.

\subsection{Evidence-Grounded Response and Why It Works}

Selected episodes are expanded into a compact buffer containing their original turns, speakers, timestamps, and available multimodal fields. If $\Gamma$ denotes this expansion, then
\begin{equation}
\hat{y}=G\!\left(q,H(q),\Gamma(\mathcal{E}(q))\right).
\end{equation}
The answer is generated from this evidence rather than from index records alone. Fundamentally, the design works by separating \emph{localization} from \emph{interpretation}: a compact structured record makes the relevant evidence addressable, while the source episode preserves the wording and context needed to reason correctly. Compatible property granularity reduces lexical mismatch, and query-type-aware coverage prevents a multi-fact question from stopping after one merely related episode.

Consider the Appendix case asking what Melanie's children like. A topical memory system can surface plausible family activities yet miss the requested child-interest facts. EnSIMem represents the request as the entity \emph{Melanie's children} and property \emph{likes}, recognizes that the answer is aggregative, and retrieves separate episodes stating their interest in nature and dinosaurs. Because the buffer contains those original statements rather than a regenerated summary, the answer model sees both the correct values and their provenance. The same reusable offline memory can serve later requests, while each online context remains short, focused, and grounded.

The same principle applies when the wording or reasoning operation changes. A request such as ``When did Caroline attend the support group?'' requires the system to preserve both the event and its source time, whereas ``What do Melanie's children like?'' requires coverage of multiple child-interest values. In the first case, a single temporally grounded episode can be sufficient; in the second, stopping after the first related episode would be an error. The planner and evidence buffer therefore work together: the former specifies what must be covered, and the latter supplies the unmodified text needed to resolve references, compare dates, and aggregate values. This is why the architecture can keep the online context compact without reducing the answer to a lossy generalized memory.

\section{Experiments}

\subsection{Evaluation on the LoCoMo Benchmark Dataset}

We report evaluation-model accuracy using the Memora-compatible protocol \citep{xia2026memora}. Because answers are open-ended, BLEU and F1 which measure lexical overlap rather than semantic correctness and do not represent a convincing measure; we therefore use an LLM-based, fixed binary evaluation protocol. LLM model-based evaluation is a scalable proxy for human assessment, while known evaluator biases motivate reporting two evaluation models separately \citep{zheng2023judging}. The model used for answer generation in both offline and online stages is GPT-4.1-mini. We report the GPT-4o-mini and Qwen3-32B evaluation-model results separately rather than averaging them. Benchmark definitions and category mappings are provided in Appendix~\ref{app:benchmarks}.

EnSIMem is strongest on single-hop and temporal questions, while open-domain questions remain the most challenging slice. Overall, GPT-4o-mini evaluates the answer and gives an overall accuracy of 90.6\%, and Qwen3-32B gives an overall accuracy of 90.0\%. We report these evaluation-model results separately; the GPT-4o-mini row remains the directly comparable result against the published Memora numbers.

\begin{table*}[t]
\caption{Evaluation: LLM-model-based accuracy on LoCoMo. Published baseline and MEMORA scores are transcribed from the Memora benchmark table; asterisks mark baselines reported there from prior work. We report the same EnSIMem answers under GPT-4o-mini and Qwen3-32B evaluation models.
\\}
\label{tab:locomo-accuracy}
\centering
\small
\begin{tabular}{lccccc}
\toprule
\textbf{Method} & \textbf{Multi-hop} & \textbf{Temporal} & \textbf{Open-domain} & \textbf{Single-hop} & \textbf{Overall} \\
\midrule
Full Context & 76.6\% & 81.9\% & 50.0\% & 88.5\% & 82.5\% \\
RAG & 55.7\% & 54.8\% & 45.8\% & 71.0\% & 63.3\% \\
HippoRAG & 39.0\% & 22.4\% & 51.0\% & 58.7\% & 47.1\% \\
Zep$^*$ & 53.7\% & 60.2\% & 43.8\% & 66.9\% & 61.6\% \\
Mem0 & 62.4\% & 66.0\% & 50.0\% & 67.7\% & 65.3\% \\
LangMem$^*$ & 71.0\% & 50.8\% & 59.0\% & 84.5\% & 73.4\% \\
Nemori$^*$ & 75.1\% & 77.6\% & 51.0\% & 84.9\% & 79.4\% \\
MEMORA (S) & 78.4\% & 85.1\% & 59.4\% & 90.0\% & 84.9\% \\
MEMORA (P) & 78.7\% & 86.6\% & 59.4\% & 91.8\% & 86.3\% \\
\midrule
EnSIMem (evaluated by GPT-4o-mini) & \textbf{88.7\%} & \underline{89.0\%} & \underline{70.8\%} & \underline{94.2\%} & \textbf{90.6\%} \\
EnSIMem (evaluated by Qwen3-32B) & \underline{83.3\%} & \textbf{89.6\%} & \textbf{72.9\%} & \textbf{94.5\%} & \underline{90.0\%} \\
\bottomrule
\end{tabular}
\end{table*}

As a protocol check, we ran the released MEMORA (P) implementation locally: its overall score was 86.23\%, versus 86.3\% reported by Memora. This close agreement supports using the published MEMORA and baseline rows for comparison with our EnSIMem results.

\subsection{Evaluation on the LongMemEval Benchmark Dataset}

We report evaluation-model accuracy using the same two-model protocol. Comparison-row provenance is provided in Appendix~\ref{app:benchmarks}.

\begin{table*}[t]
\caption{Evaluation: LLM-model-based accuracy on LongMemEval. ``SS'' denotes single-session. Published comparison rows are transcribed from Memora; EnSIMem reports the results from both evaluation models separately. Bold denotes the highest value in each column; underlining denotes the second-highest value.}
\label{tab:longmemeval-accuracy}
\centering
\small
\setlength{\tabcolsep}{2.5pt}
\renewcommand{\arraystretch}{1.08}
\resizebox{0.96\textwidth}{!}{%
\begin{tabular}{@{}lccccccc@{}}
\toprule
\textbf{Method} & \textbf{SS-pref.} & \textbf{SS-assist.} & \textbf{Temporal} & \textbf{Multi-session} & \textbf{Knowledge-update} & \textbf{SS-user} & \textbf{Average} \\
\midrule
Full Context & 16.7\% & \textbf{98.2\%} & 60.2\% & 51.1\% & 76.9\% & 85.7\% & 65.6\% \\
Nemori & \textbf{86.7\%} & 92.9\% & 72.2\% & 55.6\% & 79.5\% & 90.0\% & 74.6\% \\
MEMORA (S) & 76.7\% & 76.8\% & 84.2\% & 73.7\% & \underline{96.2\%} & \underline{97.1\%} & 83.8\% \\
MEMORA (P) & \underline{83.3\%} & 78.6\% & \textbf{89.5\%} & 78.2\% & \textbf{97.4\%} & \textbf{98.6\%} & \underline{87.4\%} \\
\midrule
EnSIMem (evaluated by GPT-4o-mini) & \underline{83.3\%} & 91.7\% & \underline{89.3\%} & \textbf{93.0\%} & \underline{96.2\%} & \textbf{98.6\%} & \textbf{92.8\%} \\
EnSIMem (evaluated by Qwen3-32B) & \textbf{86.7\%} & \underline{94.6\%} & 89.1\% & \underline{91.4\%} & 96.1\% & \textbf{98.6\%} & \textbf{92.8\%} \\
\bottomrule
\end{tabular}
}
\end{table*}

Both evaluation models reach 92.8\% overall on LongMemEval, with the largest gains over MEMORA (P) on SS-assistant and multi-session. We report the two evaluation-model results separately rather than averaging them.

\subsection{Efficiency}

We compare online latency and exclude offline construction. EnSIMem records
planning, retrieval, generation, and reasoning-buffer tokens; MEMORA values are
transcribed from its published LoCoMo results \citep{xia2026memora}, using its
Episodic (Segment) + Factual configuration for token count.

\begin{table}[t]
\caption{Mean online efficiency on LoCoMo. EnSIMem search includes
requirement-aware planning and evidence retrieval. The MEMORA token count
is reported under its published memory-context accounting.}
\label{tab:locomo-efficiency}
\centering
\small
\setlength{\tabcolsep}{5pt}
\begin{tabular}{lrrrr}
\toprule
\textbf{System} &
\textbf{Search (s)} &
\textbf{Generation (s)} &
\textbf{End-to-end (s)} &
\textbf{Context tokens} \\
\midrule
MEMORA (P) & 4.61 & N/A$^{\ddagger}$ & 5.70 & 8,499$^{\dagger}$ \\
EnSIMem & 11.9 & 2.57 & 14.4 & 7,470 \\
\bottomrule
\end{tabular}

\vspace{0.25em}
\raggedright
\footnotesize
$^{\dagger}$ The MEMORA value is transcribed from its published LoCoMo
ablation table rather than its latency table; the two systems may use
different token-accounting conventions.

$^{\ddagger}$ MEMORA does not report a separate generation-latency
breakdown in its published efficiency table.
\end{table}

EnSIMem is slower online than MEMORA (14.4 versus 5.70 seconds), with the
difference concentrated in requirement-aware planning (10.3 seconds) and
retrieval (1.59 seconds); generation takes 2.57 seconds. This cost buys
explicit property, reasoning-type, and coverage checks before the short buffer
is built. EnSIMem uses 7,470 context tokens versus MEMORA's reported 8,499;
because accounting may differ, this comparison is directional. Caching,
batching, smaller planners, and early stopping are future optimization targets.
\subsection{Ablation Study}

We conduct controlled ablations on a fixed LoCoMo subset, changing one design choice while fixing the answer model, evaluation model, and evaluation protocol. Each comparison isolates one source of the final gain rather than retuning the full pipeline.

\begin{figure*}[t]
    \centering
    \includegraphics[width=0.5\textwidth]{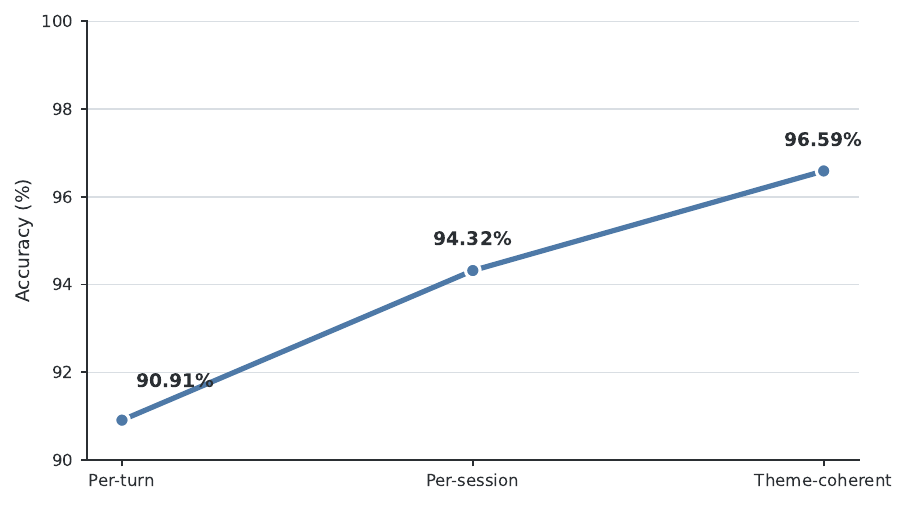}\hfill
    \includegraphics[width=0.5\textwidth]{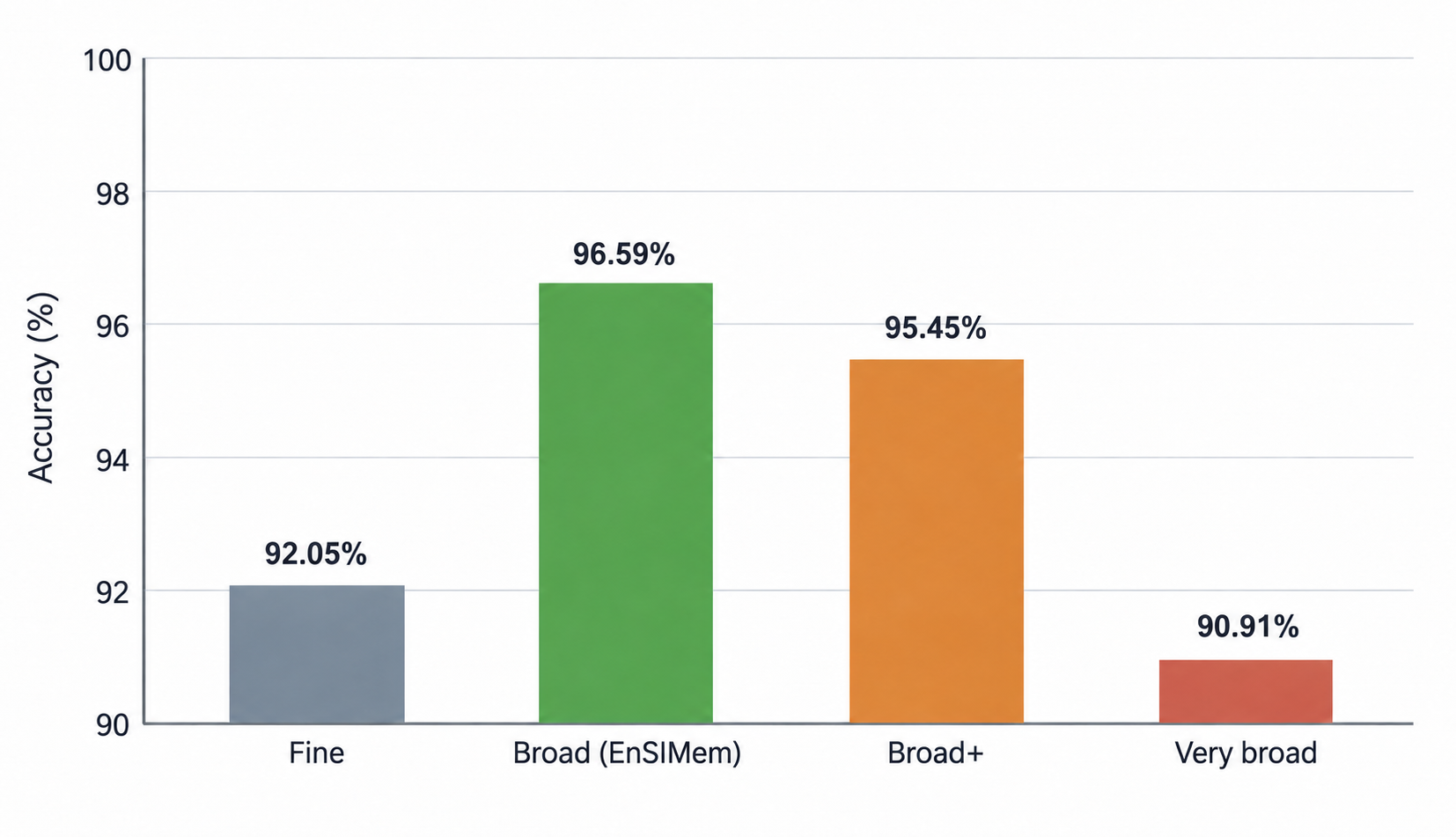}
    \caption{Ablation results on LoCoMo. The left panel compares episode
granularity, showing that theme-coherent episodes outperform per-turn and
per-session units. The right panel compares four property granularities:
Fine, Broad (EnSIMem setting), Broad+, and Very broad. Broad+ is a
slightly broader variant that merges additional action-like predicates into
the shared property \emph{activity}, while still preserving intermediate
property distinctions. Higher accuracy is better.}
    \label{fig:ablation-combined}
\end{figure*}

\paragraph{Episode granularity.} We compare theme-coherent episodes with per-turn and per-session units. This tests whether coherent middle-sized units improve the relevance--noise trade-off for an agent's reasoning buffer. The results are shown in Figure~\ref{fig:ablation-combined}. EnSIMem obtains 96.59\% accuracy, compared with 94.32\% for per-session episodes and 90.91\% for per-turn episodes. Thus, theme-coherent episodes improve accuracy by 2.27 and 5.68 percentage points over the two alternatives, respectively. This isolates memory-unit granularity: per-turn units can lose local references, whereas per-session units can admit unrelated context.
\paragraph{Property granularity.}
Using the same episodes, we compare four extraction granularities. Fine preserves
surface predicates; Broad+ is slightly broader than EnSIMem's Broad prompt and
merges additional action-like predicates into \emph{activity}; Very broad maps
more properties to \emph{fact} or broad concepts. Accuracy is 92.05\%, 96.59\%, 95.45\%, and
90.91\% for Fine, Broad (EnSIMem), Broad+, and Very broad, respectively, showing that
moderate normalization helps while excessive coarsening hurts retrieval. These
results suggest that moderate normalization improves property alignment, while
overly coarse properties remove distinctions needed for retrieval.

\paragraph{Structured matching versus dense retrieval.} Dense retrieval ranks complete episodes by semantic similarity, whereas EnSIMem first matches entity, type, and property fields and links records to source episodes. With the answer and evaluation models fixed, EnSIMem reaches 96.59\% versus 86.36\% for dense retrieval (Table~\ref{tab:ablation-retrieval-route}), showing the benefit of the structured access path. It therefore tests whether explicit index structure contributes beyond dense semantic similarity alone.

\begin{table}[t]
    \caption{Retrieval-route ablation on LoCoMo. Both variants use the same answer model and GPT-4o-mini evaluation model.}
    \label{tab:ablation-retrieval-route}
    \centering
    \small
    \begin{tabular}{lc}
        \toprule
        \textbf{Retrieval variant} & \textbf{Accuracy} \\
        \midrule
        Dense retrieval only & 86.36\% \\
        \textbf{EnSIMem: structured entity/type/property retrieval} & \textbf{96.59\%} \\
        \bottomrule
    \end{tabular}
\end{table}
\begin{figure}[t]
    \centering
    \includegraphics[width=0.7\columnwidth]{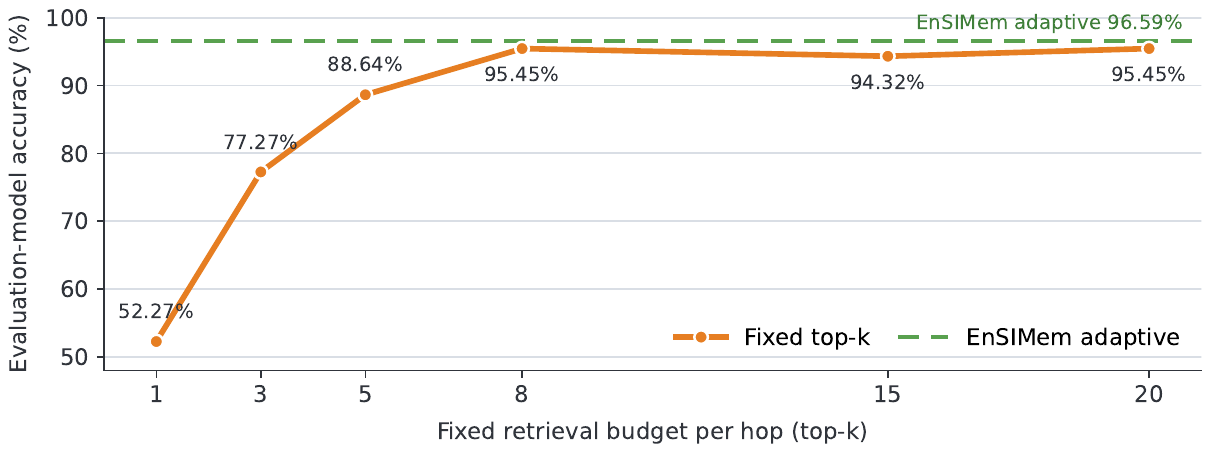}
    \caption{Fixed top-$k$ budgets versus EnSIMem's adaptive budget on LoCoMo. The dashed line shows the adaptive result. Higher accuracy is better.}
    \label{fig:ablation-topk}
\end{figure}

\paragraph{Retrieval Episode Strategy.} We compare the adaptive budget with fixed top-$k$ policies for $k\in\{1,3,5,8,15,20\}$. Fixed retrieval rises from 52.27\% at $k=1$ to 95.45\% at $k=8$ before saturating or declining, while EnSIMem reaches 96.59\% (Figure~\ref{fig:ablation-topk}). This isolates query-type-aware allocation: fixed budgets trade recall and noise uniformly, whereas the adaptive policy can spend evidence where the question requires coverage.

\section{Discussion, Limitations, and Future Work}

This study focuses on episodic conversational memory. LoCoMo and
LongMemEval evaluate conversational recall rather than procedural skill
execution or long-term semantic consolidation. The pipeline also relies on
LLM-based segmentation, property extraction, and query planning, so changes
in the underlying models may affect the intermediate memory representation.

The main practical limitation is online latency. EnSIMem is slower than
MEMORA, with most of the additional cost coming from requirement-aware
planning and evidence retrieval rather than answer generation. This cost is
intentional: the planner identifies the answer target, property granularity,
reasoning type, and required evidence coverage before constructing the
reasoning buffer. Nevertheless, repeatedly performing this analysis for every
query can be expensive for interactive applications. Future implementations
could cache plans for recurring query patterns, batch property matching,
distill planning into a smaller model, and apply confidence-aware early
stopping for simple point questions. These optimizations should reduce
latency without weakening evidence coverage or source grounding.

The same address-and-provenance principle may also extend beyond episodic
memory. In principle, semantic and procedural settings would require
extracting different information---for example, stable facts for semantic
memory or reusable action-related information for procedural memory---while
preserving links to the evidence from which it was derived. A unified agent
could then coordinate multiple memory types according to the request. We
leave the concrete representations, algorithms, and evaluation of these
extensions to future work.

\section{Conclusion}

EnSIMem reframes long-term conversational memory as a short-context reasoning
problem. Theme-coherent episodes preserve the local context needed to
interpret references, while entity--property records provide precise handles
for locating relevant evidence. Requirement-aware planning and adaptive
retrieval then select only the evidence needed for point, temporal,
compositional, and aggregation questions. The reasoning buffer contains the
original source text, provenance, and available multimodal evidence rather
than generalized summaries, allowing the answer model to reason over grounded
content without exposing it to the entire interaction history. This
evidence-preserving design provides a foundation for reliable long-term
memory and future extensions to semantic and procedural memory.

\section*{AI Use Disclosure}

In this work, we used generative AI tools to polish the writing, verify grammar errors, and implement and debug software. We also used GPT-4.1-mini as the answer-generation model in our benchmark experiments and GPT-4o-mini and Qwen3-32B as evaluation models for assessing generated answers. These model-based evaluations were conducted as part of the experimental protocol and were not used to create synthetic training data or alter the ground-truth annotations.

We did not use generative AI tools to generate synthetic datasets, provide proofs, or establish mathematical claims without author verification. All AI-assisted code was reviewed, executed, and tested by the authors. The authors independently checked the manuscript text, citations, mathematical notation, figures, experimental configurations, and reported results. We take full responsibility for the final content of this paper.

\section*{Reproducibility Statement}
The supplementary material contains the two code packages used in our experiments, \texttt{LoCoMo/} and \texttt{LongMemEval/}. The \texttt{LoCoMo/} package includes preprocessing, episode construction, property extraction, entity-index construction, requirement-aware query planning, structured and dense retrieval, answer generation, ablation experiments, and evaluation. The \texttt{LongMemEval/} package contains the corresponding benchmark preparation, inference, and evaluation scripts. Both packages include the prompts, configuration files, input/output formats, result-aggregation scripts, and execution instructions needed to reproduce the reported results. The public benchmark data are obtained from the official LoCoMo and LongMemEval releases and are not redistributed; API credentials are not included.

All experiments use GPT-4.1-mini for answer generation, GPT-4o-mini and Qwen3-32B as evaluation models, temperature 0, and random seed 42 where supported. 

\bibliography{references}

@article{liu2024lost,
  title   = {Lost in the Middle: How Language Models Use Long Contexts},
  author  = {Liu, Nelson F. and Lin, Kevin and Hewitt, John and Paranjape, Ashwin and Bevilacqua, Michele and Petroni, Fabio and Liang, Percy},
  journal = {Transactions of the Association for Computational Linguistics},
  volume  = {12},
  pages   = {157--173},
  year    = {2024},
  doi     = {10.1162/tacl_a_00638}
}

@inproceedings{wang2024loong,
  title     = {Leave No Document Behind: Benchmarking Long-Context {LLMs} with Extended Multi-Document {QA}},
  author    = {Wang, Minzheng and Chen, Longze and Fu, Cheng and Liao, Shengyi and Zhang, Xinghua and Wu, Bingli and Yu, Haiyang and Xu, Nan and Zhang, Lei and Luo, Run and Li, Yunshui and Yang, Min and Huang, Fei and Li, Yongbin},
  booktitle = {Proceedings of the 2024 Conference on Empirical Methods in Natural Language Processing},
  year      = {2024},
  doi       = {10.18653/v1/2024.emnlp-main.322}
}

@inproceedings{bertsch2025oolong,
  title     = {{Oolong}: Evaluating Long Context Reasoning and Aggregation Capabilities},
  author    = {Bertsch, Amanda and Pratapa, Adithya and Mitamura, Teruko and Neubig, Graham and Gormley, Matthew R.},
  booktitle = {Conference on Language Modeling},
  year      = {2026},
  url       = {https://colm.eventhosts.cc/Conferences/2026/AcceptedPapers}
}

@inproceedings{bai2025longbenchv2,
  title     = {{LongBench v2}: Towards Deeper Understanding and Reasoning on Realistic Long-Context Multitasks},
  author    = {Bai, Yushi and Tu, Shangqing and Zhang, Jiajie and Peng, Hao and Wang, Xiaozhi and Lv, Xin and Cao, Shulin and Xu, Jiazheng and Hou, Lei and Dong, Yuxiao and Tang, Jie and Li, Juanzi},
  booktitle = {Proceedings of the 63rd Annual Meeting of the Association for Computational Linguistics (Volume 1: Long Papers)},
  year      = {2025},
  doi       = {10.18653/v1/2025.acl-long.183}
}

@inproceedings{lewis2020rag,
  title     = {Retrieval-Augmented Generation for Knowledge-Intensive {NLP} Tasks},
  author    = {Lewis, Patrick and Perez, Ethan and Piktus, Aleksandra and Petroni, Fabio and Karpukhin, Vladimir and Goyal, Naman and K{\"u}ttler, Heinrich and Lewis, Mike and Yih, Wen-tau and Rockt{\"a}schel, Tim and Riedel, Sebastian and Kiela, Douwe},
  booktitle = {Advances in Neural Information Processing Systems},
  volume    = {33},
  pages     = {9459--9474},
  year      = {2020}
}

@inproceedings{asai2024selfrag,
  title     = {{Self-RAG}: Learning to Retrieve, Generate, and Critique through Self-Reflection},
  author    = {Asai, Akari and Wu, Zeqiu and Wang, Yizhong and Sil, Avirup and Hajishirzi, Hannaneh},
  booktitle = {International Conference on Learning Representations},
  year      = {2024}
}

@inproceedings{trivedi2023ircot,
  title     = {Interleaving Retrieval with Chain-of-Thought Reasoning for Knowledge-Intensive Multi-Step Questions},
  author    = {Trivedi, Harsh and Balasubramanian, Niranjan and Khot, Tushar and Sabharwal, Ashish},
  booktitle = {Proceedings of the 61st Annual Meeting of the Association for Computational Linguistics (Volume 1: Long Papers)},
  pages     = {10014--10037},
  year      = {2023},
  doi       = {10.18653/v1/2023.acl-long.557}
}

@inproceedings{karpukhin2020dpr,
  title     = {Dense Passage Retrieval for Open-Domain Question Answering},
  author    = {Karpukhin, Vladimir and Oguz, Barlas and Min, Sewon and Lewis, Patrick and Wu, Ledell and Edunov, Sergey and Chen, Danqi and Yih, Wen-tau},
  booktitle = {Proceedings of the 2020 Conference on Empirical Methods in Natural Language Processing},
  pages     = {6769--6781},
  year      = {2020},
  doi       = {10.18653/v1/2020.emnlp-main.550}
}

@article{izacard2022contriever,
  title   = {Unsupervised Dense Information Retrieval with Contrastive Learning},
  author  = {Izacard, Gautier and Caron, Mathilde and Hosseini, Lucas and Riedel, Sebastian and Bojanowski, Piotr and Joulin, Armand and Grave, Edouard},
  journal = {Transactions on Machine Learning Research},
  year    = {2022}
}

@inproceedings{khattab2020colbert,
  title     = {{ColBERT}: Efficient and Effective Passage Search via Contextualized Late Interaction over {BERT}},
  author    = {Khattab, Omar and Zaharia, Matei},
  booktitle = {Proceedings of the 43rd International ACM SIGIR Conference on Research and Development in Information Retrieval},
  pages     = {39--48},
  year      = {2020},
  doi       = {10.1145/3397271.3401075}
}

@article{edge2024graphrag,
  title   = {From Local to Global: A Graph {RAG} Approach to Query-Focused Summarization},
  author  = {Edge, Darren and Trinh, Ha and Cheng, Newman and Bradley, Joshua and Chao, Alex and Mody, Apurva and Truitt, Steven and Metropolitansky, Dasha and Ness, Robert Osazuwa and Larson, Jonathan},
  journal = {arXiv preprint arXiv:2404.16130},
  year    = {2024},
  doi     = {10.48550/arXiv.2404.16130}
}

@inproceedings{gutierrez2024hipporag,
  title     = {{HippoRAG}: Neurobiologically Inspired Long-Term Memory for Large Language Models},
  author    = {Jim{\'e}nez Guti{\'e}rrez, Bernal and Shu, Yiheng and Gu, Yu and Yasunaga, Michihiro and Su, Yu},
  booktitle = {Advances in Neural Information Processing Systems},
  volume    = {37},
  year      = {2024},
  doi       = {10.52202/079017-1902},
  url       = {https://proceedings.neurips.cc/paper_files/paper/2024/hash/6ddc001d07ca4f319af96a3024f6dbd1-Abstract.html}
}

@inproceedings{he2024gretriever,
  title     = {{G-Retriever}: Retrieval-Augmented Generation for Textual Graph Understanding and Question Answering},
  author    = {He, Xiaoxin and Tian, Yijun and Sun, Yifei and Chawla, Nitesh V. and Laurent, Thomas and LeCun, Yann and Bresson, Xavier and Hooi, Bryan},
  booktitle = {Advances in Neural Information Processing Systems},
  volume    = {37},
  pages     = {132876--132907},
  year      = {2024},
  doi       = {10.52202/079017-4224},
  url       = {https://proceedings.neurips.cc/paper_files/paper/2024/hash/efaf1c9726648c8ba363a5c927440529-Abstract.html}
}

@inproceedings{hu2024grag,
  title     = {{GRAG}: Graph Retrieval-Augmented Generation},
  author    = {Hu, Yuntong and Lei, Zhihan and Zhang, Zheng and Pan, Bo and Ling, Chen and Zhao, Liang},
  booktitle = {Findings of the Association for Computational Linguistics: NAACL 2025},
  pages     = {4145--4157},
  year      = {2025},
  doi       = {10.18653/v1/2025.findings-naacl.232},
  url       = {https://aclanthology.org/2025.findings-naacl.232/}
}

@article{parekh2025structure,
  title={Structure-Augmented Reasoning Generation},
  author={Parekh, Jash Rajesh and Jiang, Pengcheng and Han, Jiawei},
  journal={arXiv preprint arXiv:2506.08364},
  year={2025}
}

@article{shankar2024docetl,
  title     = {{DocETL}: Agentic Query Rewriting and Evaluation for Complex Document Processing},
  author    = {Shankar, Shreya and Chambers, Tristan and Shah, Tarak and Parameswaran, Aditya G. and Wu, Eugene},
  journal   = {Proceedings of the VLDB Endowment},
  volume    = {18},
  number    = {9},
  pages     = {3035--3048},
  year      = {2025},
  doi       = {10.14778/3746405.3746426},
  url       = {https://www.vldb.org/pvldb/vol18/p3035-shankar.pdf}
}

@article{joshi2026sliders,
  title   = {{SLIDERS}: Systematic Reviews via Automated Evidence Synthesis and Reconciliation},
  author  = {Joshi, Harshit and Shethia, Priyank and Dao, Jadelynn and Lam, Monica S.},
  journal = {arXiv preprint arXiv:2604.22294},
  year    = {2026},
  doi     = {10.48550/arXiv.2604.22294}
}

@inproceedings{du2025context,
  title     = {Context Length Alone Hurts {LLM} Performance Despite Perfect Retrieval},
  author    = {Du, Yufeng and Tian, Minyang and Ronanki, Srikanth and Rongali, Subendhu and Bodapati, Sravan Babu and Galstyan, Aram and Wells, Azton and Schwartz, Roy and Huerta, Eliu A. and Peng, Hao},
  booktitle = {Findings of the Association for Computational Linguistics: EMNLP 2025},
  pages     = {23281--23298},
  year      = {2025},
  doi       = {10.18653/v1/2025.findings-emnlp.1264},
  url       = {https://aclanthology.org/2025.findings-emnlp.1264/}
}

@inproceedings{zheng2023judging,
  title     = {Judging {LLM}-as-a-{Judge} with {MT-B}ench and {Chatbot Arena}},
  author    = {Zheng, Lianmin and Chiang, Wei-Lin and Sheng, Ying and Zhuang, Siyuan and Wu, Zhanghao and Zhuang, Yonghao and Lin, Zi and Li, Zhuohan and Li, Dacheng and Xing, Eric P. and Zhang, Hao and Gonzalez, Joseph E. and Stoica, Ion},
  booktitle = {Advances in Neural Information Processing Systems},
  volume    = {36},
  pages     = {46595--46623},
  year      = {2023},
  url       = {https://proceedings.neurips.cc/paper_files/paper/2023/hash/91f18a1287b398d378ef22505bf41832-Abstract-Datasets_and_Benchmarks.html}
}

@inproceedings{maharana2024locomo,
  title     = {Evaluating Very Long-Term Conversational Memory of {LLM} Agents},
  author    = {Maharana, Adyasha and Lee, Dong-Ho and Tulyakov, Sergey and Bansal, Mohit and Barbieri, Francesco and Fang, Yuwei},
  booktitle = {Proceedings of the 62nd Annual Meeting of the Association for Computational Linguistics (Volume 1: Long Papers)},
  pages     = {13829--13849},
  year      = {2024},
  doi       = {10.18653/v1/2024.acl-long.747}
}

@article{xia2026memora,
  title   = {Memora: A Harmonic Memory Representation Balancing Abstraction and Specificity},
  author  = {Xia, Menglin and Zhang, Xuchao and Dixit, Shantanu and Harimurugan, Paramaguru and Wang, Rujia and Ruhle, Victor and Sim, Robert and Bansal, Chetan and Rajmohan, Saravan},
  journal = {arXiv preprint arXiv:2602.03315},
  year    = {2026},
  doi     = {10.48550/arXiv.2602.03315}
}

@article{meng2026ensirag,
  title   = {{EnSI-RAG}: Entity-Structure-Indexed Retrieval-Augmented Generation for Long-Document Question Answering},
  author  = {Meng, Xuanyu and Sun, Jiashuo and Parekh, Jash Rajesh and Han, Jiawei},
  journal = {arXiv preprint arXiv:2608.21252},
  year    = {2026},
  doi     = {10.48550/arXiv.2608.21252}
}

@inproceedings{wu2024longmemeval,
  title     = {{LongMemEval}: Benchmarking Chat Assistants on Long-Term Interactive Memory},
  author    = {Wu, Di and Wang, Hongwei and Yu, Wenhao and Zhang, Yuwei and Chang, Kai-Wei and Yu, Dong},
  booktitle = {International Conference on Learning Representations},
  year      = {2025},
  url       = {https://proceedings.iclr.cc/paper_files/paper/2025/hash/d813d324dbf0598bbdc9c8e79740ed01-Abstract-Conference.html}
}

@inproceedings{park2023generative,
  title     = {Generative Agents: Interactive Simulacra of Human Behavior},
  author    = {Park, Joon Sung and O'Brien, Joseph C. and Cai, Carrie J. and Morris, Meredith Ringel and Liang, Percy and Bernstein, Michael S.},
  booktitle = {Proceedings of the 36th Annual ACM Symposium on User Interface Software and Technology},
  year      = {2023},
  doi       = {10.1145/3586183.3606763}
}

@inproceedings{packer2023memgpt,
  title     = {{MemGPT}: Towards {LLMs} as Operating Systems},
  author    = {Packer, Charles and Wooders, Sarah and Lin, Kevin and Fang, Vivian and Patil, Shishir G. and Stoica, Ion and Gonzalez, Joseph E.},
  booktitle = {International Conference on Learning Representations},
  year      = {2024},
  url       = {https://openreview.net/forum?id=0Kk142lP62}
}

@inproceedings{zhong2023memorybank,
  title     = {{MemoryBank}: Enhancing Large Language Models with Long-Term Memory},
  author    = {Zhong, Wanjun and Guo, Lianghong and Gao, Qiqi and Ye, He and Wang, Yanlin},
  booktitle = {Proceedings of the AAAI Conference on Artificial Intelligence},
  volume    = {38},
  number    = {17},
  pages     = {19724--19731},
  year      = {2024},
  doi       = {10.1609/AAAI.V38I17.29946},
  url       = {https://doi.org/10.1609/AAAI.V38I17.29946}
}

@inproceedings{shinn2023reflexion,
  title     = {Reflexion: Language Agents with Verbal Reinforcement Learning},
  author    = {Shinn, Noah and Cassano, Federico and Gopinath, Ashwin and Narasimhan, Karthik and Yao, Shunyu},
  booktitle = {Advances in Neural Information Processing Systems},
  year      = {2023}
}

@article{wang2023voyager,
  title   = {Voyager: An Open-Ended Embodied Agent with Large Language Models},
  author  = {Wang, Guanzhi and Xie, Yuqi and Jiang, Yunfan and Mandlekar, Ajay and Xiao, Chaowei and Zhu, Yuke and Fan, Linxi and Anandkumar, Anima},
  journal = {Transactions on Machine Learning Research},
  year    = {2024},
  url     = {https://openreview.net/pdf?id=ehfRiF0R3a}
}

@article{rasmussen2025zep,
  title   = {Zep: A Temporal Knowledge Graph Architecture for Agent Memory},
  author  = {Rasmussen, Preston and Paliychuk, Pavlo and Beauvais, Travis and Ryan, Jack and Chalef, Daniel},
  journal = {arXiv preprint arXiv:2501.13956},
  year    = {2025},
  doi     = {10.48550/arXiv.2501.13956}
}

@inproceedings{chhikara2025mem0,
  title     = {{Mem0}: Building Production-Ready {AI} Agents with Scalable Long-Term Memory},
  author    = {Chhikara, Prateek and Khant, Dev and Aryan, Saket and Singh, Taranjeet and Yadav, Deshraj},
  booktitle = {European Conference on Artificial Intelligence},
  pages     = {2993--3000},
  year      = {2025},
  doi       = {10.3233/FAIA251160},
  url       = {https://doi.org/10.3233/FAIA251160}
}

@inproceedings{xu2025amem,
  title     = {{A-M}EM: Agentic Memory for {LLM} Agents},
  author    = {Xu, Wujiang and Liang, Zujie and Mei, Kai and Gao, Hang and Tan, Juntao and Zhang, Yongfeng},
  booktitle = {Advances in Neural Information Processing Systems},
  volume    = {38},
  pages     = {20004--20031},
  year      = {2025},
  doi       = {10.52202/085713-0593},
  url       = {https://proceedings.neurips.cc/paper_files/paper/2025/hash/19909c36f51abc4856b4560aff3d36d6-Abstract-Conference.html}
}

@inproceedings{kang2025memoryos,
  title     = {Memory {OS} of {AI} Agent},
  author    = {Kang, Jiazheng and Ji, Mingming and Zhao, Zhe and Bai, Ting},
  booktitle = {Proceedings of the 2025 Conference on Empirical Methods in Natural Language Processing},
  pages     = {25961--25970},
  year      = {2025},
  doi       = {10.18653/v1/2025.emnlp-main.1318},
  url       = {https://aclanthology.org/2025.emnlp-main.1318/}
}

@article{huang2026agentmemorysurvey,
  title   = {A Survey of Agent Memory in the Second Half: Towards Self-Evolving and Long-Horizon Agents},
  author  = {Huang, Wei-Chieh and Zhang, Weizhi and Liang, Yueqing and Bei, Yuanchen and Chen, Yankai and Feng, Tao and Pan, Xinyu and Tan, Zhen and Wang, Yu and Wei, Tianxin and Wu, Shanglin and Xu, Ruiyao and Yang, Liangwei and Yang, Rui and Yang, Wooseong and Yeh, Chin-Yuan and Zhang, Hanrong and Zhang, Haozhen and Zhu, Siqi and Zou, Henry Peng and Zhao, Wanjia and Wang, Song and Xu, Wujiang and Ke, Zixuan and Hui, Zheng and Li, Dawei and Wu, Yaozu and He, Langzhou and Wang, Chen and Xu, Xiongxiao and Huang, Baixiang and Tan, Juntao and Heinecke, Shelby and Wang, Huan and Xiong, Caiming and Metwally, Ahmed and Yan, Jun and Lee, Chen-Yu and Zeng, Hanqing and Xia, Yinglong and Wei, Xiaokai and Payani, Ali and Wang, Yu and Ma, Haitong and Wang, Wenya and Wang, Chenguang and Zhang, Yu and Wang, Xin Eric and Zhang, Yongfeng and You, Jiaxuan and Tong, Hanghang and Luo, Xiao and Liu, Xue and Sun, Yizhou and Wang, Wei and McAuley, Julian and Zou, James and Han, Jiawei and Yu, Philip S. and Shu, Kai},
  journal = {Transactions on Machine Learning Research},
  year    = {2026},
  issn    = {2835-8856},
  url     = {https://openreview.net/forum?id=XycbogUAeJ},
  note    = {Survey Certification}
}

@inproceedings{yang2026plugmem,
  title   = {{PlugMem}: A Task-Agnostic Plugin Memory Module for {LLM} Agents},
  author  = {Yang, Ke and Chen, Zixi and He, Xuan and Jiang, Jize and Galley, Michel and Wang, Chenglong and Gao, Jianfeng and Han, Jiawei and Zhai, ChengXiang},
  booktitle = {Proceedings of the 43rd International Conference on Machine Learning},
  year      = {2026},
  url       = {https://icml.cc/virtual/2026/poster/64446}
}

@inproceedings{ouyang2026reasoningbank,
  title     = {{ReasoningBank}: Scaling Agent Self-Evolving with Reasoning Memory},
  author    = {Ouyang, Siru and Yan, Jun and Hsu, I-Hung and Chen, Yanfei and Jiang, Ke and Wang, Zifeng and Han, Rujun and Le, Long T. and Daruki, Samira and Tang, Xiangru and Tirumalashetty, Vishy and Lee, George and Rofouei, Mahsan and Lin, Hangfei and Han, Jiawei and Lee, Chen-Yu and Pfister, Tomas},
  booktitle = {International Conference on Learning Representations},
  year      = {2026},
  url       = {https://openreview.net/forum?id=jL7fwchScm}
}

@inproceedings{ouyang2026skillos,
  title     = {{SkillOS}: Learning Skill Curation for Self-Evolving Agents},
  author    = {Ouyang, Siru and Yan, Jun and Chen, Yanfei and Han, Rujun and Wang, Zifeng and Mishra, Bhavana Dalvi and Meng, Rui and Li, Chun-Liang and Jiao, Yizhu and Zha, Kaiwen and Shen, Maohao and Tirumalashetty, Vishy and Lee, George and Han, Jiawei and Pfister, Tomas and Lee, Chen-Yu},
  booktitle = {Advances in Neural Information Processing Systems},
  year      = {2026},
  url       = {https://arxiv.org/abs/2605.06614}
}

@inproceedings{yu2026agemem,
  title     = {Agentic Memory: Learning Unified Long-Term and Short-Term Memory Management for Large Language Model Agents},
  author    = {Yu, Yi and Yao, Liuyi and Xie, Yuexiang and Tan, Qingquan and Feng, Jiaqi and Li, Yaliang and Wu, Libing},
  booktitle = {Proceedings of the 64th Annual Meeting of the Association for Computational Linguistics (Volume 1: Long Papers)},
  pages     = {21457--21483},
  year      = {2026},
  doi       = {10.18653/v1/2026.acl-long.981}
}

@inproceedings{ma2026nemori,
  title     = {What Deserves Memory: Adaptive Memory Distillation for {LLM} Agents},
  author    = {Ma, Wenquan and Nan, Jiayan and Wu, WenLong},
  booktitle = {Proceedings of the 64th Annual Meeting of the Association for Computational Linguistics (Volume 1: Long Papers)},
  pages     = {34789--34812},
  year      = {2026},
  doi       = {10.18653/v1/2026.acl-long.1607}
}

@inproceedings{bei2026memgallery,
  title     = {{Mem-Gallery}: Benchmarking Multimodal Long-Term Conversational Memory for {MLLM} Agents},
  author    = {Bei, Yuanchen and Wei, Tianxin and Ning, Xuying and Zhao, Yanjun and Liu, Zhining and Lin, Xiao and Zhu, Yada and Hamann, Hendrik and He, Jingrui and Tong, Hanghang},
  booktitle = {Proceedings of the 64th Annual Meeting of the Association for Computational Linguistics (Volume 1: Long Papers)},
  pages     = {40750--40784},
  year      = {2026},
  doi       = {10.18653/v1/2026.acl-long.1892},
  url       = {https://aclanthology.org/2026.acl-long.1892/}
}

@article{ma2026memprobe,
  title   = {{MemAudit}: Auditing Long-Term Agent Memory via Hidden User-State Recovery},
  author  = {Ma, Enze and Zhou, Yufan and Huang, Wei-Chieh and Yang, Jie and Ma, Huanhuan and Wang, Zixuan and Li, Chengze and Miao, Chunyu and Yu, Philip S. and Wang, Zhen},
  journal = {arXiv preprint arXiv:2606.24595},
  year    = {2026},
  doi     = {10.48550/arXiv.2606.24595}
}

@article{belikova2026procedural,
  title   = {Managing Procedural Memory in {LLM} Agents: Control, Adaptation, and Evaluation},
  author  = {Belikova, Julia and Parchiev, Rauf and Egorov, Evgeny and Davydenko, Grigorii and Gusev, Gleb and Savchenko, Andrey and Makarenko, Maksim},
  journal = {arXiv preprint arXiv:2606.23127},
  year    = {2026},
  doi     = {10.48550/arXiv.2606.23127}
}

@article{song2026skillops,
  title   = {{SkillOps}: Managing {LLM} Agent Skill Libraries as Self-Maintaining Software Ecosystems},
  author  = {Pu, Hongji and Song, Xinyuan and Zhao, Liang},
  journal = {arXiv preprint arXiv:2605.13716},
  year    = {2026},
  doi     = {10.48550/arXiv.2605.13716}
}

@article{xu2026agentskills,
  title   = {Agent Skills for Large Language Models: Architecture, Acquisition, Security, and the Path Forward},
  author  = {Xu, Renjun and Yan, Yang},
  journal = {arXiv preprint arXiv:2602.12430},
  year    = {2026},
  doi     = {10.48550/arXiv.2602.12430}
}

@misc{langchain2025langmem,
  title        = {{LangMem SDK} for Agent Long-Term Memory},
  author       = {{LangChain}},
  year         = {2025},
  howpublished = {\url{https://www.langchain.com/blog/langmem-sdk-launch}}
}
\bibliographystyle{iclr2027_conference}

\clearpage
\onecolumn
\appendix
\section{Appendix}\label{app:benchmarks}

\subsection{Benchmark overview}

\begin{samepage}
\refstepcounter{table}\label{tab:benchmark-overview}
\noindent\textbf{Table~\thetable.} Benchmark scope, question families, and evaluation protocol used in this paper.\par
\vspace{0.35em}
\footnotesize
\setlength{\tabcolsep}{3pt}
\renewcommand{\arraystretch}{1.16}
\noindent\begin{tabular}{@{}p{0.18\textwidth}p{0.39\textwidth}p{0.36\textwidth}@{}}
\toprule
\textbf{Benchmark} & \textbf{Scope and question families} & \textbf{Protocol and role in this paper} \\
\midrule
\textbf{LoCoMo} & \textbf{Scope:} 10 long, multi-session conversations with speaker turns, timestamps, and optional image observations. \textbf{Question families:} multi-hop, temporal, open-domain, and single-hop questions testing entity tracking, paraphrase alignment, temporal grounding, multimodal evidence use, and aggregation. & \textbf{Protocol:} Evaluation-model scoring on the same generated answers using GPT-4o-mini and Qwen3-32B, reported separately. \textbf{Result:} 90.6\% and 90.0\% overall, respectively. Published comparison rows are transcribed from Memora \citep{maharana2024locomo,xia2026memora}. \\
\noalign{\smallskip}
\textbf{LongMemEval} & \textbf{Scope:} interactive long-term memory tasks pairing a conversation history with a later user request. \textbf{Question families:} single-session preference, single-session assistant, temporal, multi-session, knowledge update, and single-session user questions. & \textbf{Protocol:} the same two evaluation-model procedure, with GPT-4o-mini and Qwen3-32B reported separately. \textbf{Result:} 92.8\% overall for each evaluation model after rounding. Full Context, Nemori, and MEMORA values follow the published Memora table \citep{wu2024longmemeval,xia2026memora}. \\
\bottomrule
\end{tabular}
\end{samepage}

\newpage
\subsection{Baseline and method overview}

\begin{samepage}
\refstepcounter{table}\label{tab:baseline-overview}
\noindent\textbf{Table~\thetable.} Baseline and method comparison used in the benchmark tables.\par
\vspace{0.35em}
\footnotesize
\setlength{\tabcolsep}{3pt}
\renewcommand{\arraystretch}{1.12}
\noindent\begin{tabular}{@{}p{0.18\textwidth}p{0.39\textwidth}p{0.36\textwidth}@{}}
\toprule
\textbf{Method} & \textbf{Representation and access} & \textbf{Comparison role} \\
\midrule
\textbf{Full Context} & Complete interaction history is passed to the answer model; there is no persistent selection structure. & Upper reference that minimizes retrieval misses, but exposes the model to irrelevant context and high input cost. \\
\noalign{\smallskip}
\textbf{RAG} & Anonymous text chunks retrieved by dense semantic similarity. & Standard retrieval baseline; it does not explicitly address the relevant entity, property, provenance, or reasoning type \citep{lewis2020rag,karpukhin2020dpr}. \\
\noalign{\smallskip}
\textbf{HippoRAG} & Graph-inspired associative memory with propagation across related information. & Tests whether graph retrieval alone matches explicit entity-property addressing for multi-hop recall \citep{gutierrez2024hipporag}. \\
\noalign{\smallskip}
\textbf{Zep} & Temporally aware knowledge graph with evolving facts and relations. & Tests explicit temporal organization; it does not use EnSIMem's source episodes or requirement-specific evidence budgets \citep{rasmussen2025zep}. \\
\noalign{\smallskip}
\textbf{Mem0} & Consolidated facts and preferences, with graph variants for relations. & Production-oriented persistent memory; consolidation can remove local dialogue context \citep{chhikara2025mem0}. \\
\noalign{\smallskip}
\textbf{Nemori} & Persistent conversational-memory representation. & Established memory baseline; we report the published comparison values rather than reimplementing or retuning it \citep{ma2026nemori,xia2026memora}. \\
\noalign{\smallskip}
\textbf{MEMORA (S)} & Harmonic entries combine an abstraction, a richer value, and cue indices. & Sparse-retrieval baseline testing the abstraction-specificity trade-off \citep{xia2026memora}. \\
\noalign{\smallskip}
\textbf{MEMORA (P)} & The same harmonic representation with an adaptive policy retriever. & Strongest published comparator and closest policy-driven baseline \citep{xia2026memora}. \\
\noalign{\smallskip}
\bottomrule
\end{tabular}
\end{samepage}

\newpage
\section{Case Studies}

We present three representative LoCoMo cases in which the evaluation model labels EnSIMem correct while labeling Memora incorrect. The cases are selected to expose three distinct mechanisms: cross-episode evidence coverage, temporal grounding, and fine-grained entity-property localization. Each table records the question, the required operation, the memory representation, the retrieved evidence, the generated responses, and the specific failure mode of the competing system.

\begin{samepage}
\refstepcounter{table}\label{tab:case-study-kids}
\noindent\textbf{Table~\thetable.} Case study 1: cross-episode evidence coverage for a multi-fact question.\par
\vspace{0.3em}
\small
\setlength{\tabcolsep}{4pt}
\renewcommand{\arraystretch}{1.10}
\noindent\begin{tabular}{@{}p{0.22\textwidth}p{0.76\textwidth}@{}}
\toprule
\textbf{Case component} & \textbf{Details} \\
\midrule
\textbf{Question} & \textit{What do Melanie's kids like?} (query b0008; category 1). The question refers to the children as a group and asks for their salient interests, not for one isolated activity from a single conversation. \\
\midrule
\textbf{Required operation} & The system must resolve the subject \textit{Melanie's kids}, collect answer-bearing facts across sessions, and avoid replacing the requested interests with a generic list of family activities. The gold answer is \textit{dinosaurs, nature}. \\
\midrule
\textbf{Memora representation} & Memora exposes family-related memories as broad activity summaries. These memories contain camping, painting, beach, hiking, and pottery, but they do not provide an explicit address for the entity \textit{Melanie's kids} together with the property \textit{likes} and its answer-bearing values. \\
\midrule
\textbf{Memora response} & ``Melanie's kids like camping, painting, beach, hiking, pottery.'' \caseincorrect. The response is topically related, but it omits both benchmark-targeted facts: the children love nature and are interested in dinosaurs. \\
\midrule
\textbf{Why Memora fails} & Memora's broad summary conflates nearby family activities with the specific values requested by the question. The retrieval result is not empty; rather, it fails to preserve the right evidence with sufficient priority. This is a coverage-and-specificity failure caused by semantic aggregation without a structured child-interest address. \\
\midrule
\textbf{EnSIMem query structure} & The query is represented as an entity-property requirement over Melanie's children and their interests. Because the answer is aggregative, the planner keeps expanding evidence until the relevant child-interest facts are covered instead of stopping after one semantically similar episode. \\
\midrule
\textbf{EnSIMem representation} & Theme-coherent episodes preserve the local dialogue and keep each extracted entity-property record linked to its source turns. This prevents facts about Melanie's own activities from being confused with facts about her children. \\
\midrule
\textbf{EnSIMem retrieval} & EnSIMem retrieves \texttt{conv-26::session\_4::episode\_002} (D4:8), where Melanie says that the two younger children love nature, and \texttt{conv-26::session\_6::episode\_003} (D6:6), where the children are excited about a dinosaur exhibit. The two episodes jointly cover the gold answer. \\
\midrule
\textbf{Evidence trace} & D4:8 states that the children ``love nature'' after a family camping trip. D6:6 states that they were excited about a dinosaur exhibit and enjoyed learning about animals. These are direct, dialogue-grounded supports rather than inferred preferences. \\
\midrule
\textbf{EnSIMem response} & EnSIMem identifies the children's interest in nature and dinosaurs and retains the supporting episode references. \casecorrect. \\
\midrule
\textbf{Why EnSIMem succeeds} & Structured matching first identifies the correct entity and property, while adaptive expansion supplies the second episode needed for complete coverage. The answer buffer therefore contains the answer-bearing facts instead of only a broad collection of related activities. \\
\midrule
\textbf{Key difference} & Memora returns a generic semantic activity summary. EnSIMem uses an addressable entity-property index and query-type-aware evidence coverage to aggregate the right facts across multiple episodes before generation. \\
\bottomrule
\end{tabular}
\end{samepage}

\newpage
\begin{samepage}
\refstepcounter{table}\label{tab:case-study-book-time}
\noindent\textbf{Table~\thetable.} Case study 2: temporal grounding from a relative-time expression.\par
\vspace{0.3em}
\small
\setlength{\tabcolsep}{4pt}
\renewcommand{\arraystretch}{1.10}
\noindent\begin{tabular}{@{}p{0.22\textwidth}p{0.76\textwidth}@{}}
\toprule
\textbf{Case component} & \textbf{Details} \\
\midrule
\textbf{Question} & \textit{When did Melanie read the book ``Nothing is Impossible''?} (query b0039; category 2). The question requires resolving a relative temporal expression rather than retrieving a date stated as an absolute year. \\
\midrule
\textbf{Required operation} & Link the book-reading event to its temporal qualifier, interpret ``last year'' relative to the dated conversation, and return the normalized year. The gold answer is \textit{2022}. \\
\midrule
\textbf{Memora representation} & Memora's memory store contains many book and reading memories, but it does not maintain a directly addressable relation between Melanie, the reading event, the book reference, and the relative-time phrase in the source episode. \\
\midrule
\textbf{Memora response} & ``The memories do not mention Melanie reading the book `nothing is impossible'.'' \caseincorrect. The source statement is present in the conversation, but it is not surfaced as an answerable memory. \\
\midrule
\textbf{Why Memora fails} & The failure is relational rather than purely lexical. The question names the book explicitly, while the source says ``this book'' and places the event inside a dated conversation. A generic memory lookup therefore treats the title as absent and does not connect the event to its relative-time qualifier. \\
\midrule
\textbf{EnSIMem query structure} & The planner creates a temporal requirement for Melanie's reading activity and the referenced book. It marks the relative-time expression as answer-bearing, so retrieval must preserve both the source timestamp and the phrase ``last year.'' \\
\midrule
\textbf{EnSIMem representation} & The entity-property record links Melanie to the reading event and retains the value-level temporal qualifier. The original episode remains available for interpreting the reference ``this book'' and for checking the date used in normalization. \\
\midrule
\textbf{EnSIMem retrieval} & EnSIMem retrieves \texttt{conv-26::session\_7::episode\_002}, specifically D7:8: ``This book I read last year reminds me to always pursue my dreams.'' The episode is dated 12 July 2023. \\
\midrule
\textbf{Evidence trace} & The phrase ``last year'' is directly grounded in D7:8. Applying the episode date as the reference point gives 2022, which is the benchmark answer. \\
\midrule
\textbf{EnSIMem response} & EnSIMem answers that Melanie read the book ``last year'' and normalizes the relative expression to 2022. \casecorrect. \\
\midrule
\textbf{Why EnSIMem succeeds} & The index preserves the event, entity, temporal qualifier, and source timestamp as a connected retrieval path. Query-specific planning then performs the small amount of temporal normalization needed after retrieval. \\
\midrule
\textbf{Key difference} & Memora requires a directly matching memory statement and reports no evidence. EnSIMem preserves temporal relations as searchable structure and uses the source episode to resolve relative time without losing provenance. \\
\bottomrule
\end{tabular}
\end{samepage}

\newpage
\begin{samepage}
\refstepcounter{table}\label{tab:case-study-art}
\noindent\textbf{Table~\thetable.} Case study 3: entity-property specificity for multimodal art evidence.\par
\vspace{0.3em}
\small
\setlength{\tabcolsep}{4pt}
\renewcommand{\arraystretch}{1.10}
\noindent\begin{tabular}{@{}p{0.22\textwidth}p{0.76\textwidth}@{}}
\toprule
\textbf{Case component} & \textbf{Details} \\
\midrule
\textbf{Question} & \textit{What kind of art does Caroline make?} (query b0017; category 1). The question asks for the relevant artistic form, not merely whether Caroline engages in visual art. \\
\midrule
\textbf{Required operation} & Identify Caroline as the entity, distinguish the artistic-form property from the broad activity of making art, and use both textual and image-grounded evidence. The gold answer is \textit{abstract art}. \\
\midrule
\textbf{Memora representation} & Memora stores broad activity-level descriptions such as painting and drawing. These summaries do not expose a sufficiently specific property for the artistic form, and the associated image descriptions are not retained as an explicit answer handle. \\
\midrule
\textbf{Memora response} & ``Painting and drawing.'' \caseincorrect. The response is related to Caroline's activities but is too coarse to answer the requested subtype. \\
\midrule
\textbf{Why Memora fails} & The representation collapses several distinct records into one generic art activity. It therefore cannot distinguish the requested style from the broader fact that Caroline paints or draws, and it does not use the abstract-landscape image description as a retrieval cue. \\
\midrule
\textbf{EnSIMem query structure} & The planner addresses the entity Caroline and requests the property describing the kind or form of her art. The requirement is specific enough to prefer an abstract-art record over generic records about painting, drawing, or art as emotional expression. \\
\midrule
\textbf{EnSIMem representation} & EnSIMem indexes Caroline's art-related records at an intermediate property granularity and links each record to its source episode, original dialogue, timestamp, and multimodal fields. This creates a direct path from Caroline to the artistic-form property. \\
\midrule
\textbf{EnSIMem retrieval} & The system retrieves \texttt{conv-26::session\_9::episode\_003} (D9:14), whose image caption describes an abstract landscape, and \texttt{conv-26::session\_11::episode\_003} (D11:8 and D11:12), which preserve Caroline's paintings about inclusivity, identity, and self-acceptance. \\
\midrule
\textbf{Evidence trace} & D9:14 supplies the image-grounded abstract-landscape cue. D11:8 and D11:12 confirm that the artwork belongs to Caroline and connect the visual artifacts to her stated artistic practice rather than to Melanie's art activities. \\
\midrule
\textbf{EnSIMem response} & The retrieved evidence supports the more specific answer \textit{abstract art}. The evaluation model labels the EnSIMem answer \casecorrect. \\
\midrule
\textbf{Why EnSIMem succeeds} & The entity-property index retains the distinction between a broad activity and a requested subtype, while provenance-preserving episode expansion supplies the surrounding dialogue and image evidence needed to interpret the artwork. \\
\midrule
\textbf{Key difference} & Memora retrieves a generic activity category. EnSIMem preserves the entity, property granularity, source episode, and image-grounded evidence together, allowing the agent to answer at the specificity required by the benchmark. \\
\bottomrule
\end{tabular}
\end{samepage}

\newpage
\section{Evaluation Model Rationale}

Our evaluation follows the answer-scoring protocol used by Memora and its underlying Mem0 evaluation implementation \citep{xia2026memora}. This choice is appropriate for long-term agent memory because the benchmark answers are open-ended: a correct answer may paraphrase the reference, use a relative temporal expression, or include additional explanation without matching the reference string word for word.

\paragraph{LoCoMo decision rule.} For every query, the evaluation program passes the question, the gold answer, and the generated answer to an evaluation model. The Memora/Mem0-compatible prompt asks whether the generated answer is semantically consistent with the gold answer and requests a binary \texttt{CORRECT} or \texttt{WRONG} label in JSON. The implementation maps \texttt{CORRECT} to one and \texttt{WRONG} to zero, then aggregates these decisions overall and by LoCoMo category. Category 5 is excluded according to the Memora-compatible scoring policy. Thus, the reported LoCoMo number measures answer-level semantic correctness rather than token overlap.

\paragraph{LongMemEval decision rule.} LongMemEval uses the corresponding Memora-style task-specific checker. The evaluation program again separates answer generation from evaluation, but selects a prompt according to the question type: preference questions use a personalization rubric, knowledge-update questions check the updated answer, temporal questions include their temporal tolerance, and unanswerable items check whether the model correctly abstains. The evaluator receives the question, reference answer or rubric, and generated response, and returns \texttt{yes} or \texttt{no}; the implementation maps \texttt{yes} to correct and aggregates accuracy by question type and overall. This preserves the benchmark's intended semantics while keeping the evaluation procedure independent of the retrieval pipeline.

\paragraph{Why this is preferable to lexical metrics.} BLEU and F1 primarily reward shared surface tokens. They can penalize a valid paraphrase, fail to recognize equivalent date formats, and provide little evidence that the answer identifies the requested entity, property, or event. The evaluation-model prompt instead exposes the complete question--gold--prediction triple and explicitly allows semantically equivalent wording and time expressions. This makes the protocol better aligned with the actual objective of an agent-memory system: recovering the right fact or event for the user's request.

\paragraph{Reproducibility and model dependence.} LoCoMo evaluation uses temperature zero, a fixed seed of 42, and a fixed Memora-compatible prompt; the LongMemEval checker uses temperature zero and its fixed task-specific prompts. The same protocol and aggregation code are used across methods within each benchmark. Because model-based evaluation can still reflect evaluator-specific preferences, we report GPT-4o-mini and Qwen3-32B results separately rather than treating their arithmetic mean as a new ground truth. Agreement in the qualitative trends across the two evaluation models provides a more transparent robustness check while preserving the identity of each evaluation model.

\paragraph{Relation to the benchmark comparison.} Using the Memora-compatible protocol keeps our LoCoMo comparison on the same answer-level scale as the published Memora rows. It also makes the source of every score explicit: the answer model generates the response, the evaluation model makes the binary semantic decision, and the reported accuracy is the fraction of correct decisions. This separation avoids conflating answer generation with evaluation and allows future work to replace or add evaluation models without changing the stored answers or retrieval pipeline.

\clearpage
\section{Evaluation-Model Consistency}\label{app:evaluation-consistency}

We additionally examine whether the reported scores are stable across the two evaluation models.
For each benchmark category, let $g_c$ and $q_c$ denote the accuracy obtained with GPT-4o-mini and
Qwen3-32B, respectively, and define the signed difference
$\Delta_c = q_c-g_c$ in percentage points (pp). We report the mean signed difference, mean
absolute difference (MAE), the population standard deviation of the category differences, the
largest absolute category difference, the gap between the two overall scores, and the Pearson
correlation across category accuracies. The analysis uses the unrounded category-level results
underlying the main tables before their three-significant-digit display; it does not treat the
arithmetic mean of the two evaluation models as a new score.

The two evaluation models therefore give highly similar aggregate conclusions. On LoCoMo, the
overall scores differ by only $0.610$ pp (90.6\% versus 90.0\%), and the category-level MAE is
$2.11$ pp. The larger LoCoMo discrepancy is concentrated in multi-hop questions, where Qwen3-32B
is $5.46$ pp lower; the remaining category gaps are at most $2.09$ pp. On LongMemEval, the overall
gap is only $0.030$ pp (92.8\% versus 92.8\% in three significant figures), with a smaller category-level MAE of $1.37$ pp.
The largest LongMemEval difference is $3.37$ pp on single-session preference questions, while
temporal reasoning, knowledge update, and single-session user questions differ by at most $0.250$
pp. The high category-level correlations ($r=0.946$ for LoCoMo and $r=0.936$ for LongMemEval),
together with the small absolute gaps, indicate that the headline findings are consistent across
the two evaluation models even though individual categories can show model-specific sensitivity.

\begin{table}[t]
\centering
\caption{Consistency of GPT-4o-mini and Qwen3-32B evaluation-model scores. $\Delta$ is defined as Qwen3-32B minus GPT-4o-mini; all difference columns are in percentage points. The correlation is computed across benchmark categories, excluding the overall column.}
\label{tab:evaluation-consistency}
\small
\setlength{\tabcolsep}{4pt}
\renewcommand{\arraystretch}{1.08}
\begin{tabular}{@{}lrrrrrrr@{}}
\toprule
\textbf{Benchmark} & $m$ & $\overline{\Delta}$ & $\mathrm{MAE}$ & $\mathrm{SD}(\Delta)$ & $\max|\Delta|$ & $|\Delta_{\mathrm{overall}}|$ & $r$ \\
\midrule
LoCoMo & 4 & $-0.620$ & $2.11$ & $2.88$ & $5.46$ & $0.610$ & $0.946$ \\
LongMemEval & 6 & $+0.750$ & $1.37$ & $1.79$ & $3.37$ & $0.030$ & $0.936$ \\
\bottomrule
\end{tabular}
\end{table}

This is an aggregate stability analysis rather than a claim of perfect per-query agreement. A
paired reliability analysis would require retaining both binary labels for every identical query.
With those labels, one could additionally report exact agreement, Cohen's $\kappa$, a paired
McNemar test, and bootstrap confidence intervals. We do not infer those quantities from category
totals alone; instead, we report the two evaluation-model results separately and use the agreement
of their aggregate trends as the robustness check.

\clearpage
\section{Prompts}\label{app:prompts}

This appendix records the core prompts used by EnSIMem. The text inside each box is the prompt
template passed to the corresponding LLM stage; braces denote runtime substitutions such as a
conversation, an episode, a query, or a retrieval plan. We show the prompts that determine the
memory representation, query decomposition, evidence-grounded answering, and evaluation. Repair
and validation prompts are invoked only when a model returns malformed JSON and follow the same
schemas shown here. Structured matching, dense fallback, evidence-budget expansion, and score
aggregation are deterministic operations and do not invoke an LLM.
The corresponding source files are \texttt{LoCoMo/prompts.py},
\texttt{LoCoMo/08\_memora\_llm\_judge.py}, and
\texttt{LongMemEval/memora\_evaluation.py}; the runtime substitutions shown below are made by
the pipeline scripts.

\subsection{Offline memory construction}

\paragraph{Theme-coherent episode partitioning.} The partition prompt creates contiguous evidence
units before any entity-property records are extracted. It preserves dialogue order and keeps image
metadata attached to the turn that introduced it.

\begin{promptbox}[blue]{Theme-Coherent Episode Partitioning: System Prompt}
You partition a complete dialogue session into contiguous theme-coherent memory episodes. These boundaries define immutable evidence units. Return one JSON object only.
\end{promptbox}

\begin{promptbox}[blue]{Theme-Coherent Episode Partitioning: User Prompt}
Partition this session.

A memory episode is one contiguous span mainly focused on one coherent entity, event, goal, or bounded topic. Use theme_type entity, event, or topic. Split only on a genuine semantic shift; keep greetings, acknowledgements, clarification questions, and short follow-ups with the material they support.

Structural rules:
- Cover every supplied dialogue turn exactly once and in order, without gaps or overlaps.
- Do not split merely because the speaker changes.
- Avoid one-turn fragments unless a turn clearly introduces a separate durable theme.
- Keep text, image captions, image-retrieval descriptions, and image URLs attached to their DIA.
- Copy start_dia_id and end_dia_id exactly.

Return JSON only:
{"segments": [{"start_dia_id": "D1:1", "end_dia_id": "D1:5", "theme": "concise canonical theme", "theme_type": "entity|event|topic", "theme_description": "one-sentence scope", "boundary_reason": "session_start or semantic shift"}]}

Conversation: {conversation_id}
Session: {session_id}
Observed at: {observed_at}
Turns:
{session_text}
\end{promptbox}

\paragraph{Entity-property index extraction.} This prompt is query-independent: it extracts atomic
records once and keeps the source episode as the only answer evidence. The property policy is
deliberately intermediate-grained, while values and source DIA identifiers retain concrete detail.

\begin{promptbox}[teal]{Entity-Property Indexing: System Prompt}
You are the high-recall, precision-preserving information-extraction stage of an episodic memory system. Build a query-independent entity-structured index from one complete theme episode. The episode remains the only answer evidence; records are navigation handles. Follow the minimum-sufficient predicate policy and return one valid JSON object only.
\end{promptbox}

\begin{promptbox}[teal]{Entity-Property Indexing: User Prompt}
Extract an exhaustive set of atomic records from the episode below.

Read every dialogue turn and every displayed metadata field. Image captions and image-retrieval descriptions are auxiliary textual evidence; an image URL is opaque provenance and must never be used for outside lookup.

Extraction requirements:
- Scan each DIA independently and emit a record for every explicit factual clause.
- Resolve I/my/me using the speaker on that turn and cite exact evidence_dia_ids.
- Preserve the speaker-to-entity relation for explicit first-person actions and emit explicit inverse relations only when entailed.
- Use one entity and one minimum-sufficient reusable property. Keep distinctive predicates such as research, travel, attend, read, paint, checkup, and support; do not collapse them all into activity or concatenate the subject into the property.
- Put concrete source detail in value. Preserve relative time, modality, polarity, conditions, and image metadata.
- Do not infer facts from world knowledge, episode adjacency, or an image URL.

Return one object only:
{"records": [{"entity": "...", "entity_type": "person|organization|place|object|event|activity|concept|other", "property": "short predicate", "value": "explicit value or empty", "condition_property": "... or empty", "condition_value": "... or empty", "property_kind": "aspect|relation", "modality": "observed|planned|desired|hypothetical|negated|uncertain", "valid_time": "... or empty", "source": "speaker", "evidence_dia_ids": ["D1:1"], "projection_kind": "direct|inverse|state_projection|event_access", "confidence": 0.0}]}

Episode metadata:
Episode ID: {episode_id}
Conversation: {conversation_id}
Session: {session_id}
Observed at: {observed_at}
Episode theme: {episode_theme} ({episode_theme_type})

Episode turns:
{episode_text}
\end{promptbox}

\subsection{Online retrieval and response generation}

\paragraph{Requirement-aware query planning.} The planner converts a question into explicit
requirements and an evidence-seeking hop graph. It uses the same schema as the index, marks
aggregation questions as \texttt{all\_matching}, and leaves unknown answer values empty.

\begin{promptbox}[violet]{Requirement-Aware Query Planning: System Prompt}
You are the exhaustive query-decomposition stage of an episodic memory system. Convert the question into searchable requirements and an evidence-seeking hop graph using the same [entity][entity_type][property:value][condition:value] schema as the memory index. Preserve every meaningful discriminator, but never guess an answer. Do not answer the question. Return exactly one valid JSON object.
\end{promptbox}

\begin{promptbox}[violet]{Requirement-Aware Query Planning: User Prompt}
Build an exhaustive retrieval plan for the question below.

Planning rules:
- Include the named entity, the relationship that disambiguates the target, the requested property, and all explicit time, quantity, and location constraints.
- Use minimum-sufficient properties aligned with the index; do not invent subject-prefixed or compound labels.
- Keep unknown answer values empty. Use only information stated in the question or a declared bridge.
- Add separate hops for independently useful discriminators. For example, a daughter's birthday requires relationship=daughter, property=birthday, and property=time.
- Use reasoning_type=aggregation or comparison and retrieval_scope=all_matching for counts, exhaustive lists, frequencies, intervals, and first/second questions. Otherwise use point retrieval.
- Preserve relative time and explicit conditions. Do not answer the question in this stage.

Return one object only:
{"answer_target": {"type": "time|value|entity|location|count|list|boolean|likelihood|explanation", "description": "..."}, "reasoning_type": "direct|aggregation|comparison|temporal|causal|counterfactual|commonsense_inference|multi_hop|unanswerable", "retrieval_scope": "point|all_matching", "required_properties": [{"entity": "...", "entity_type": "...", "broad_property": "...", "property_text": "...", "value": "...", "role": "entity|relation|answer_property|time|constraint"}], "hops": [{"hop_id": "h1", "purpose": "...", "anchor": {"entity": "", "entity_type": "", "property": "", "value": "", "condition_property": "", "condition_value": ""}, "depends_on": [], "bridge_request": ""}]}

Observed index property vocabulary:
{index_property_vocabulary}

Question: {question}
\end{promptbox}

\paragraph{Evidence-grounded answer generation.} The answer model receives complete retrieved
episodes, not only the structured records. The structured records and hop plan are navigation aids;
the final answer must be supported by the original dialogue and attached image metadata.

\begin{promptbox}[red]{Evidence-Grounded Answer Generation: System Prompt}
Answer only from the supplied complete original conversation episodes. The structured indexes and hop plan are navigation aids, not evidence. Read evidence from every hop, combine facts when the plan is multi-hop, and keep entities correctly bound. Do not use outside knowledge. Treat captions and retrieval descriptions as textual evidence attached to their exact DIA; treat image URLs as opaque provenance.

For list, count, frequency, interval, and first/second questions, inventory every qualifying event across all supplied episodes before answering. For temporal questions, bind the requested event first and preserve approximate source wording. If the evidence does not establish the answer, output exactly: Unknown. Return only the shortest sufficient answer without explanation.
\end{promptbox}

\begin{promptbox}[red]{Evidence-Grounded Answer Generation: User Prompt}
Question: {question}

Retrieval plan:
{plan}

Complete original episodes retrieved across all hops:
{episodes}

Answer:
\end{promptbox}

\subsection{Evaluation prompts}

The benchmark evaluation is separate from answer generation. The evaluator receives the question,
the benchmark gold answer (or rubric), and the generated answer. It does not receive the retrieval
plan or hidden retrieval diagnostics. LoCoMo uses the Memora-compatible binary prompt below;
LongMemEval selects the corresponding task-specific variant according to the question type.

\begin{promptbox}[orange]{LoCoMo Evaluation Model: Single User Prompt}
Your task is to label an answer to a question as CORRECT or WRONG. You will be given (1) a question, (2) a gold answer, and (3) a generated answer. Be generous with grading: a longer answer is correct when it touches the same topic as the gold answer, and a time answer is correct when it refers to the same date or period even if the format differs.

Question: {question}
Gold answer: {gold_answer}
Generated answer: {generated_answer}

Return only a JSON object with the key "label" and the value CORRECT or WRONG.
\end{promptbox}

\begin{promptbox}[orange]{LongMemEval Evaluation Model: Task-Specific User Prompt}
I will give you a question, a reference answer or rubric, and a model response. Answer yes if the response is semantically correct and no otherwise. The response may use equivalent wording. For preference questions, check whether it satisfies the personalization rubric. For knowledge-update questions, accept the updated answer even if older information is also present. For temporal-reasoning questions, do not penalize an off-by-one error when the benchmark asks for a number of days, weeks, or months. For unanswerable questions, answer yes only when the response correctly identifies that the requested information is not given.

Question: {question}
Reference answer or rubric: {answer}
Model response: {hypothesis}

Answer yes or no only.
\end{promptbox}

The LoCoMo implementation maps \texttt{CORRECT} to one and \texttt{WRONG} to zero; the
LongMemEval implementation maps \texttt{yes} to one and \texttt{no} to zero. Both protocols use
temperature zero. The LoCoMo evaluator also fixes the seed to 42, while the LongMemEval checker
uses top-$p=1$. Category-wise and overall accuracies are computed from these binary decisions.

\end{document}